\documentclass[letterpaper]{article} 

\usepackage[preprint]{aaai2027} 

\usepackage[hyphens]{url} 
\usepackage{graphicx} 
\usepackage{natbib} 
\usepackage{caption} 
\usepackage{amsmath}
\usepackage{amssymb}
\usepackage{amsfonts}
\usepackage{amsthm}
\usepackage{nicefrac}

\usepackage{booktabs}

\graphicspath{{./}}
\DeclareGraphicsExtensions{.pdf,.png,.jpeg}

\newcommand{\E}{\mathbb{E}}

\newcommand{\Lsmooth}{\mathcal{L}_{\mathrm{smooth}}}

\newcommand{\Lcdl}{\mathcal{L}_{\mathrm{CDL\text{-}RL}}}

\newcommand{\ve}{\mathbf{e}}

\title{Spectral Initialization and Scheduled Graph Smoothness\\for Uncertain Knowledge Graph Completion}

\author{
Md Abrar Jahin\textsuperscript{\rm 1},
Taufikur Rahman Fuad\textsuperscript{\rm 2},
Jay Pujara\textsuperscript{\rm 1},
Craig A. Knoblock\textsuperscript{\rm 1}
}

\affiliations{
\textsuperscript{\rm 1}University of Southern California \quad
\textsuperscript{\rm 2}Islamic University of Technology\\[2pt]
\texttt{jahin@usc.edu}, \texttt{taufikur@iut-dhaka.edu},\\
\texttt{jpujara@usc.edu}, \texttt{knoblock@isi.edu}
}

\begin{document}

\maketitle

\begin{abstract}
  Uncertain knowledge graphs (UKGs) extend knowledge graphs by assigning each triple a continuous confidence score. Since most possible triples lack observed confidences, recent methods rely on semi-supervised learning to generate pseudo-labels. These methods initialize entity embeddings without using the confidence-weighted graph, discarding its global community and hub structure. We introduce \textbf{QUEST}, which adds no trainable parameters to the standard confidence-distribution learning pipeline. First, QUEST initializes entity embeddings using the smallest non-trivial eigenvectors of the confidence-weighted graph Laplacian, incorporating community and hub structure before training. Second, QUEST applies an unbiased mini-batch Dirichlet energy regularizer to enforce early-stage structural consistency. On two UKG datasets, QUEST improves confidence prediction and link prediction on six of eight metric–dataset pairs over prior methods and matches the previous best on the remaining two, while removing the instability spike observed on dense graphs. These results indicate that spectral structural priors combined with a graph Dirichlet energy regularizer improve accuracy, training stability, and checkpoint reliability in UKG completion.
\end{abstract}

\section{Introduction}
\label{sec:intro}
Real-world Knowledge Graphs (KGs) are fundamental for data representation and automated reasoning, but their construction naturally introduces noise and varying degrees of certainty. To capture this ambiguity, Uncertain Knowledge Graphs (UKGs) annotate each relational triple with a continuous confidence score. Modeling uncertainty in KGs is necessary as embeddings are systematically uncalibrated, which reduces their reliability in downstream reasoning tasks~\cite{tabacofProbabilityCalibrationKnowledge2020}. UKG completion aims to infer missing facts with their confidence levels. Because most possible triples lack observed confidences, recent works formulate UKG completion as a semi-supervised learning problem~\cite{passleaf2021, sscdl2025}.

Foundational UKG completion methods adapt deterministic scoring functions to regress or bound confidence scores. UKGE~\cite{ukge2019} maps bilinear scores to confidence values through hand-designed transformations, while BEUrRE~\cite{beurre2021} models uncertainty through geometric box-intersection volumes. To address label sparsity, PASSLEAF~\cite{passleaf2021} introduces a pool-based semi-supervised framework, and UPGAT~\cite{upgat2023} extends graph attention to UKGs. The state-of-the-art framework, ssCDL~\cite{sscdl2025}, models confidence as a distribution over discrete bins and introduces a meta-learning stage that generates pseudo-labels for unobserved triples. Despite these advances, existing methods randomly initialize entity embeddings and process triples locally. By lifting discrete entities into a continuous vector space without structural priors, they discard the global community and hub topology.

This structural disconnect affects optimization stability and representational quality. In deterministic KGs, spectral methods and graph-smoothness regularization have long been used to encode community structure~\cite{ng2001spectral, kipfSemiSupervisedClassificationGraph2017, zhouDirichletEnergyConstrained2021}. However, these ideas have not been transferred to the uncertain setting. Recent studies on self-training further show that topology-agnostic pseudo-labeling propagates errors through the graph~\cite{wangDeepInsightsNoisy2023, liuCanPseudoLabelBe2026}. In UKG completion, meta-learning modules assign pseudo-confidence labels to uniformly corrupted entity pairs. Without a structural prior, the smoothness objective pulls entity pairs together while the ranking gradient for corrupted triples pushes them apart. On dense graphs, this gradient conflict induces training instability and causes sharp error spikes.

We address this disconnect with QUEST, a drop-in extension of ssCDL that introduces two parameter-free interventions. First, QUEST initializes entity embeddings using the $k$ smallest non-trivial eigenvectors of the confidence-weighted graph Laplacian. This spectral initialization places entities from the same high-confidence communities near each other and centers high-degree hubs, grounding the embeddings in global topology before training begins. Second, QUEST imposes an unbiased mini-batch graph-smoothness regularizer derived from Dirichlet energy~\cite{zhouDirichletEnergyConstrained2021, liSmoothnessGeneralOptimization2024}. We find that this regularizer conflicts with meta-learned self-training on dense graphs. QUEST therefore uses a scheduling strategy that deactivates the smoothness penalty exactly when pseudo-labeling begins, yielding to meta-learned supervision while preserving the spectral prior.

Our main contributions are:
\textbf{(i)} We identify a structural disconnect in UKG completion and analyze the gradient conflict between graph-smoothness regularization and uniform entity corruption.
\textbf{(ii)} We propose a spectral initialization method with no additional trainable parameters that embeds global community and hub structure from the confidence-weighted graph Laplacian into the initial entity representations.
\textbf{(iii)} We formulate an unbiased mini-batch graph Dirichlet energy regularizer that enforces topological smoothness during early-stage confidence distribution learning, preserving structural integrity before self-training.

\begin{figure*}[t!]
  \centering
  \includegraphics[width=0.75\textwidth]{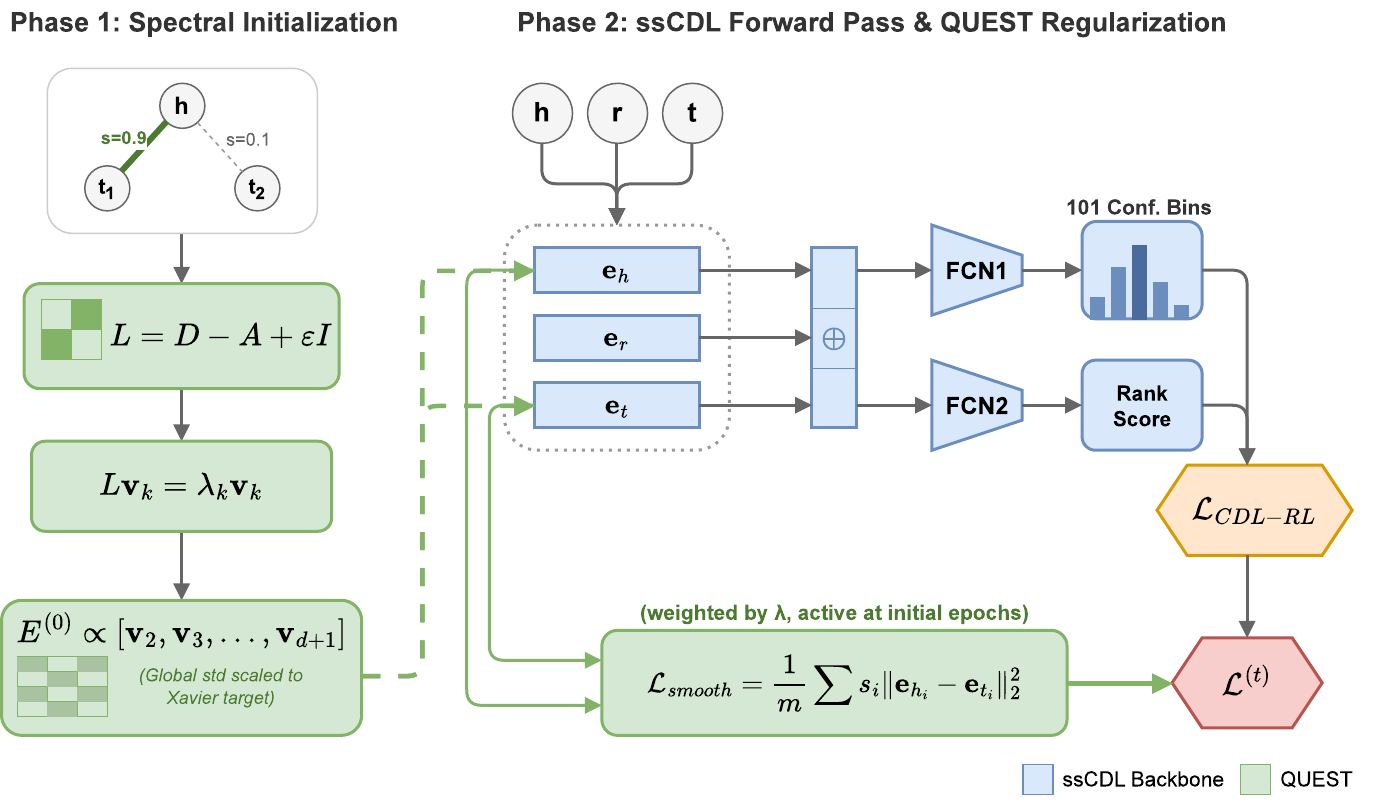}
  \caption{Overall architecture of QUEST}
  \label{fig:overview}
\end{figure*}

\section{Related Work}

\textbf{Uncertain knowledge graph completion. }
Foundational UKG embedding models adapt deterministic scoring functions. UKGE~\cite{ukge2019} introduces UKGE$_{\text{logi}}$ and UKGE$_{\text{rect}}$ that map bilinear scores to confidence values through logistic and rectified linear transformations, respectively. These hand-designed mappings produce point estimates and cannot capture distributional uncertainty. BEUrRE~\cite{beurre2021} represents entities as axis-aligned hyperrectangles and relations as affine maps, estimating confidence through box-intersection volumes. These computations scale with embedding dimension, increasing complexity. PASSLEAF~\cite{passleaf2021} integrates multiple scoring functions with a semi-supervised sample pool to mitigate label sparsity, yet still treats triples independently without exploiting graph topology. UPGAT~\cite{upgat2023} constructs an attention graph from predicted Knowledge Graph Embedding (KGE) scores rather than observed confidence weights, weakening the supervisory signal from annotated certainty values. UKGsE~\cite{ukgse2022} combines embedding-based inference with an approximate logical framework but optimizes entity vectors without graph-smoothness constraints, allowing entities in the same high-confidence cluster to drift apart.

Probabilistic entity representations have also been proposed for multi-hop KG reasoning~\cite{choudharyProbabilisticEntityRepresentation2021}. Universal orthogonal parameterization further generalizes KGE models~\cite{liGeneralizingKnowledgeGraph2024}, while spectral clustering has been used to select anchor entities before propagation~\cite{liangClusteringThenPropagation2024}. KGE models are systematically uncalibrated, reducing reliability in downstream tasks~\cite{tabacofProbabilityCalibrationKnowledge2020}. Neighborhood-intervention consistency has been proposed as a post-hoc confidence measure for KGE predictions~\cite{wangNeighborhoodInterventionConsistency2021}. ssCDL~\cite{sscdl2025} models confidence as a full distribution and couples it with PCDG, a MAML-style~\cite{finn2017model} meta-training stage that generates pseudo-confidence labels. Despite state-of-the-art results, ssCDL, like prior UKG methods, randomly initializes entity embeddings and uses graph structure only through triple corruption, discarding the global community and hub topology encoded in the confidence-weighted graph.

\textbf{Semi-supervised and pseudo-confidence learning for UKGs. }
UKG completion is inherently semi-supervised because most triples lack observed confidences. Self-training and pseudo-labeling are used in graph representation learning to propagate supervision to unlabeled nodes~\cite{liuCanPseudoLabelBe2026, wangDeepInsightsNoisy2023}, while meta-learning approaches have been developed for few-shot semi-supervised learning~\cite{dingMetaPropagationNetworks2022}. In the UKG setting, ssCDL's PCDG stage~\cite{sscdl2025} is the first to integrate meta-learning into pseudo-confidence generation. A dedicated meta-network produces soft targets for uniformly corrupted entity pairs and updates them through a one-step meta-gradient.

Noisy pseudo-labels propagate errors through graph and degrade performance as label noise increases~\cite{wangDeepInsightsNoisy2023}. Topology-agnostic pseudo-label selection underperforms structure-aware alternatives~\cite{liuCanPseudoLabelBe2026}. Structure-sensitive node selection outperforms uniform strategies in self-training~\cite{wangBANGSGametheoreticNode2025}. PCDG suffers from the same issue: uniformly corrupting entities ignores structural distance in the confidence-weighted graph. When a corrupted entity lies far from the original, the resulting pseudo-label opposes graph smoothness and introduces gradient interference.

\textbf{Structural priors and spectral methods for KGE. }
In deterministic KGs, structure-aware architectures like CompGCN~\cite{vashishth2020composition} and relational GCNs propagate information along typed edges, but they assume binary truth values and require graph access at inference time. Spectral methods provide a parameter-free alternative. Graph Laplacian eigenvectors encode community and hub structure and have been used for spectral clustering~\cite{ng2001spectral}, positional encodings, and GNN pre-training~\cite{daiLearningLaplacianEigenvectors2025}. Graph convolution has been derived as a first-order approximation of spectral filtering through a normalized Laplacian~\cite{kipfSemiSupervisedClassificationGraph2017}, though this idea remained confined to GNNs rather than embedding-based completion.

Recent work exploits spectral structure through attention mechanisms~\cite{chenStructureAwareTransformerGraph2022}, graph-smoothness unrolling~\cite{doInterpretableLightweightTransformer2024}, and graph-based structural priors~\cite{saniExploitingGraphBasedStructural2026}, but these approaches add trainable components and complexity. In contrast, QUEST initializes embeddings from the $k$ smallest non-trivial eigenvectors of the \emph{confidence-weighted} Laplacian, placing entities from the same communities near each other and positioning bridging hubs at structural centers without additional parameters or pre-training.

\textbf{Graph smoothness and regularization. }
Regularizing embeddings to be smooth over a graph is classically formulated as minimizing the Dirichlet energy. In the GNN literature, Dirichlet energy has been used to prevent over-smoothing in deep networks~\cite{zhouDirichletEnergyConstrained2021}, support joint graph learning and data imputation~\cite{xuJointFeatureDifferentiable2023, zhangDataImputationPerspective2024}, and extend Laplacian regularization to negative weights~\cite{liSmoothnessGeneralOptimization2024}. Label-smoothness regularization has also been applied to KG-aware recommendation~\cite{wangKnowledgeawareGraphNeural2019}, though these methods assume binary interaction labels and do not directly transfer to continuous UKG confidence values. Spectral regularization can also interfere with adaptive graph training when not carefully balanced~\cite{youGraphDomainAdaptation2023}.

Existing methods keep the regularizer active throughout training. QUEST differs in three ways: it regularizes entity embeddings through confidence-weighted Laplacian without GNN, estimates regularizer through unbiased mini-batch sampling, and deactivates when PCDG begins. This scheduling avoids gradient conflict between pseudo-label training and graph smoothness on dense graphs.

\section{Background and Preliminaries}
\label{sec:prelim}

\subsection{Uncertain Knowledge Graphs}
\label{sec:ukgs}

A UKG is a tuple $\mathcal{G} = (\mathcal{V}, \mathcal{R}, \mathcal{E})$, where $\mathcal{V}$ is a set of entities, $\mathcal{R}$ a set of relation types, and $\mathcal{E} \subseteq \mathcal{V} \times \mathcal{R} \times \mathcal{V} \times (0,1]$ is a set of \emph{confidence-annotated triples} $(h, r, t, s)$. Each triple states that relation $r$ holds between head entity $h$ and tail entity $t$ with confidence $s \in (0, 1]$.

We study two tasks. Confidence prediction takes a query triple $(h, r, t)$ and predicts its confidence $s$; performance is measured by Mean Squared Error (MSE) and Mean Absolute Error (MAE). Link prediction takes a query pair $(h, r, \cdot)$ and ranks candidate tail entities $t' \in \mathcal{V}$; performance is measured by confidence-weighted mean reciprocal rank (WMRR) and Hits@1 in the filtered setting~\cite{ukge2019}. We let $n = |\mathcal{V}|$, $d$ denote the embedding dimension, and $B = 101$ the number of confidence bins.

\paragraph{ssCDL backbone.}
QUEST builds on ssCDL~\cite{sscdl2025}, which jointly learns confidence prediction and link prediction over uncertain triples. For each triple $(h,r,t)$, ssCDL learns entity and relation embeddings and constructs the feature vector
\[
  \phi(h,r,t) = [\mathbf{e}_h \,\|\, \mathbf{e}_r \,\|\, \mathbf{e}_t] \in \mathbb{R}^{3d}.
\]
A confidence distribution learner maps this feature vector to a categorical distribution over $B=101$ confidence bins in $[0,1]$, trained against a truncated Gaussian target centered at the observed confidence score $s$. The final predicted confidence $\hat{s}(h,r,t)$ is obtained as the expected bin value and normalized to $[0.1,1.0]$.

In parallel, ssCDL uses a second scoring network for link prediction, optimized with a confidence-weighted margin ranking loss over negatively corrupted triples. The confidence prediction loss and ranking loss are combined using Kendall uncertainty weighting~\cite{kendall2018multi}, yielding the joint objective
\[
  \mathcal{L}_{\mathrm{cdl}}
  =
  \frac{\mathcal{L}_{\mathrm{cp}}}{2\sigma_1^2}
  +\log\sigma_1
  +
  \frac{\mathcal{L}_{\mathrm{rl}}}{20\sigma_2^2}
  +\log\sigma_2,
\]
where $\sigma_1$ and $\sigma_2$ are learnable task uncertainty parameters.

\section{QUEST}
\label{sec:method}

We propose \textbf{QUEST}, an extension of ssCDL~\cite{sscdl2025} with two parameter-free additions targeting a weakness in UKG completion: the loss of structure once entities are lifted into embeddings. First, entity embeddings are initialized from the $k$ smallest non-trivial eigenvectors of the \emph{confidence-weighted graph Laplacian} $L = D - A$, where $A_{ij} = \sum_{r} s_{ijr}$ and $D_{ii} = \sum_j A_{ij}$. This encodes community structure and hub topology at initialization. Second, QUEST adds a graph-smoothness regularizer $\Lsmooth$ to the CDL-RL objective, penalizing high-confidence neighbors that drift apart early in training. A stability analysis shows this regularizer conflicts with PCDG on dense graphs. QUEST therefore deactivates $\Lsmooth$ once PCDG begins at epoch~30, leaving spectral initialization as the structural prior. Figure~\ref{fig:overview} summarizes the pipeline.

\subsection{Confidence-Weighted Graph Laplacian}
\label{sec:laplacian}

We construct a confidence-weighted undirected graph over $\mathcal{V}$ from training triples. For each ordered pair $(i, j) \in \mathcal{V}^2$, the edge weight aggregates the confidence of triples linking $i, j$ in either direction:
\begin{equation}
  A_{ij} \;=\; \sum_{\substack{r:\,(i,r,j,s)\in\mathcal{E} \\\phantom{r:\,}\text{or}\;(j,r,i,s)\in\mathcal{E}}} s
  \label{eq:adj}
\end{equation}
Thus, $A$ is symmetric. Treating directed KG edges as undirected is standard in structural KGE methods~\cite{vashishth2020composition}. The degree matrix is $D = \mathrm{diag}(A\mathbf{1})$, and the unnormalized graph Laplacian is
\begin{equation}
  L \;=\; D - A
  \label{eq:laplacian}
\end{equation}

Because $A$ is symmetric and non-negative, $L$ is symmetric positive
semi-definite with real eigenvalues
$0 = \lambda_1 \leq \lambda_2 \leq \cdots \leq \lambda_n$.
The smallest eigenvalue $\lambda_1 = 0$ corresponds to the constant
eigenvector $\mathbf{1}/\sqrt{n}$ and is discarded. The remaining
eigenvectors $\mathbf{v}_2, \ldots, \mathbf{v}_n$, ordered by increasing
eigenvalue, form the graph spectral basis: small eigenvalues correspond
to smooth community-scale signals, while large eigenvalues correspond
to local high-frequency variations.

\subsection{Spectral Entity Initialization}
\label{sec:spectral}

\textbf{Algorithm. }
Let $d$ be the embedding dimension. We compute the $k=d$ smallest non-trivial eigenvectors of the Tikhonov-regularized Laplacian
\begin{equation}
  L_\varepsilon \;=\; L + \varepsilon I, \qquad \varepsilon = 10^{-5}
  \label{eq:lap_reg}
\end{equation}
using the ARPACK shift-invert method (shift $\sigma_0 = 0$, tolerance $10^{-4}$, maximum $2000$ Lanczos iterations). The perturbation $\varepsilon I$ removes exact singularity at $\sigma_0 = 0$, improving numerical stability. Denote the resulting eigenpairs as $(\lambda_2, \mathbf{v}_2), \ldots, (\lambda_{d+1}, \mathbf{v}_{d+1})$. Small eigenvalues $\lambda_i \approx 0$ indicate entities co-occurring in the same dense community. We assemble the spectral coordinate matrix
\begin{equation}
  U = \bigl[\mathbf{v}_2 \;\big|\; \mathbf{v}_3 \;\big|\; \cdots
  \;\big|\; \mathbf{v}_{d+1}\bigr]
  \;\in\; \mathbb{R}^{n \times d}
  \label{eq:spectral_matrix}
\end{equation}
and rescale $U$ to target standard deviation
$\tau = \sqrt{2\,/\,(n + d)}$:
\begin{equation}
  U \;\leftarrow\; \frac{\tau}{\mathrm{std}(U) + \varepsilon}\,U
  \label{eq:xavier_scale}
\end{equation}
The entity embedding table is initialized as $E^{(0)} = U$. Relation embeddings retain Xavier uniform initialization because relations do not appear in the Laplacian.

\textbf{Interpretation. }
Spectral initialization places entities in the embedding space induced by normalized spectral clustering~\cite{ng2001spectral}: entities within the same high-confidence community receive similar coordinates, while high-degree hubs bridging multiple communities occupy structurally central positions. 

\textbf{Complexity. }
Building $A$ takes $O(|\mathcal{E}|)$ time. The shift-invert ARPACK call computes the $d$ smallest eigenvectors in $O(d \cdot \mathrm{nnz}(L))$ time per Lanczos step, and typically $O(d \cdot |\mathcal{E}|)$ overall for sparse real-world graphs, where $\mathrm{nnz}(L)$ denotes the number of non-zeros. This is a one-time pre-training cost with no effect on per-epoch training or inference.

\subsection{Graph-Smoothness Regularizer}
\label{sec:graphreg}

\textbf{Formulation. }
The graph Dirichlet energy of the entity embedding matrix $E$ under the confidence-weighted graph is
$\mathrm{tr}(E^\top L E) = \sum_{(i,j)} A_{ij}\|\ve_i - \ve_j\|^2$.
We add a normalized version of this energy to the CDL-RL objective:
\begin{equation}
  \Lsmooth \;=\;
  \frac{1}{m}
  \sum_{i=1}^{m} s_i\,\|\ve_{h_i} - \ve_{t_i}\|^2
  \label{eq:lsmooth}
\end{equation}
where $(h_i, r_i, t_i, s_i)_{i=1}^{m}$ is a mini-batch of
$m = \min(2048, |\mathcal{E}|)$ edges sampled uniformly with replacement from $\mathcal{E}$ at each training step. Eq.~\eqref{eq:lsmooth} is an unbiased Monte Carlo estimate of the full Dirichlet energy $(1/|\mathcal{E}|)\,\mathrm{tr}(E^\top L E)$. Mini-batch sampling keeps the per-step overhead at $O(m \cdot d)$, negligible relative to the CDL-RL forward pass.

\textbf{Augmented objective. }
During the pre-PCDG phase ($t < 30$), the total training loss becomes
\begin{equation}
  \mathcal{L}^{(t)} \;=\; \Lcdl + \lambda\,\Lsmooth,
  \qquad t < 30
  \label{eq:augmented_loss}
\end{equation}
with $\lambda = 0.01$. Minimizing $\Lsmooth$ encourages entities connected by high-confidence edges to maintain similar embeddings, preserving the structural signal from spectral initialization during early training.

\subsection{PCDG-Aware Scheduling}
\label{sec:pcdg}

\textbf{Empirical observation.}
Leaving $\Lsmooth$ active after epoch~30 destabilizes training on CN15k. In a controlled experiment, MSE increased from $0.035$ to $0.198$ between epochs~29 and~39 with $\Lsmooth$ active. No comparable instability appeared on NL27k. The difference correlates with graph density: CN15k contains $\times\,2.5$ more edges per entity than NL27k, giving $\Lsmooth$ a stronger gradient signal.

\textbf{Mechanistic explanation.}
PCDG constructs pseudo-labels by replacing the head or tail entity in a training triple with a uniformly sampled entity. These synthetic entities are often distant from the original entity in the confidence-weighted graph. If $\Lsmooth$ remains active, it pulls distant entity embeddings together, opposing the CDL-RL gradient that assigns the synthetic entity a new confidence score. On dense graphs, the large number of high-confidence edges amplifies the $\Lsmooth$ gradient and produces persistent gradient conflict, causing the observed MSE spike.

\textbf{Scheduling rule.}
We deactivate $\Lsmooth$ when PCDG begins its first meta-update:
\begin{equation}
  \lambda^{(t)} \;=\;
  \begin{cases}
    \lambda & t < 30, \\
    0       & t \geq 30
  \end{cases}
  \label{eq:schedule}
\end{equation}
The cutoff at epoch~30 follows the fixed PCDG activation schedule; it is not tuned on the test set. After epoch~30, spectral initialization remains the persistent QUEST component. The structural prior in $E^{(0)}$ continues to shape optimization without further overhead. 

\begin{table*}[t!]
  \centering
  \small
  \begin{tabular}{l l l l l}
    \toprule
    \textbf{Dataset} & \multicolumn{2}{c}{\textbf{NL27k}} & \multicolumn{2}{c}{\textbf{CN15k}} \\
    \cmidrule(lr){2-3} \cmidrule(lr){4-5}
    \textbf{Metric} & MSE & MAE & MSE & MAE \\
    \midrule
    UKGE$_{logi}$            & 0.029 & 0.060 & 0.246 & 0.409 \\
    UKGE$_{rect}$            & 0.033 & 0.071 & 0.202 & 0.364 \\
    BEUrRE                   & 0.089 & 0.222 & 0.117 & 0.283 \\
    PASSLEAF$_{DistMult}$    & 0.023 & 0.051 & 0.216 & 0.379 \\
    PASSLEAF$_{ComplEx}$     & 0.024 & 0.052 & 0.231 & 0.400 \\
    PASSLEAF$_{RotatE}$      & 0.019 & 0.063 & 0.094 & 0.248 \\
    UKGsE                    & 0.122 & 0.271 & 0.103 & 0.256 \\
    UPGAT                    & 0.029 & 0.101 & 0.149 & 0.308 \\
    ssCDL                    & \textbf{0.009} & 0.042 & 0.034 & \textbf{0.116} \\
    \midrule
    \textbf{QUEST$_{Full}$}                   & \textbf{0.009}$_{\pm\text{0.0001}}$ & \textbf{0.041}$_{\pm\text{0.0009}}$ & \underline{0.031}$_{\pm\text{0.0002}}$ & \underline{0.118}$_{\pm\text{0.0005}}$ \\
    \textbf{QUEST}$_{\text{w/o spectral init}}$ & \underline{0.010}$_{\pm\text{0.0001}}$ & 0.044$_{\pm\text{0.0005}}$ & \underline{0.031}$_{\pm\text{0.0001}}$ & \textbf{0.116}$_{\pm\text{0.0000}}$ \\
    \textbf{QUEST}$_{\text{w/o graph-reg}}$   & \textbf{0.009}$_{\pm\text{0.0001}}$ & \underline{0.041}$_{\pm\text{0.0002}}$ & \textbf{0.030}$_{\pm\text{0.0001}}$ & \underline{0.118}$_{\pm\text{0.0004}}$ \\
    \midrule
    $\Delta$ & \textcolor{darkgray}{- - -} & \textcolor{teal}{$+2.4\%$} & \textcolor{teal}{$+11.8\%$} & \textcolor{darkgray}{- - -} \\
    \bottomrule
  \end{tabular}
  
  \caption{Comparison of QUEST and baselines on NL27k and CN15k for Confidence Prediction. Best results are in \textbf{bold}, second-best \underline{underlined}. $\Delta$ denotes percentage change relative to ssCDL; \textcolor{teal}{green} indicates improvement and \textcolor{darkgray}{- - -} indicates changes within $\pm1\%$.}
  \label{cp_result}
\end{table*}

\begin{table*}[t!]
  \centering
  \small
  \begin{tabular}{l l l l l}
    \toprule
    Dataset & \multicolumn{2}{c}{NL27k} & \multicolumn{2}{c}{CN15k} \\
    \cmidrule(lr){2-3} \cmidrule(lr){4-5}
    Metric & WMRR & Hits@1 & WMRR & Hits@1 \\
    \midrule
    UKGE$_{logi}$            & 0.593 & 0.462 & 0.118 & 0.072 \\
    UKGE$_{rect}$            & 0.580 & 0.452 & 0.127 & 0.060 \\
    BEUrRE                   & 0.272 & 0.117 & 0.138 & 0.039 \\
    PASSLEAF$_{DistMult}$    & 0.676 & 0.553 & 0.170 & 0.078 \\
    PASSLEAF$_{ComplEx}$     & 0.708 & 0.586 & 0.196 & 0.086 \\
    PASSLEAF$_{RotatE}$      & 0.715 & 0.580 & 0.137 & 0.037 \\
    UKGsE                    & 0.064 & 0.031 & 0.012 & 0.002 \\
    UPGAT                    & 0.658 & 0.530 & 0.165 & 0.078 \\
    ssCDL                    & 0.727 & 0.636 & 0.207 & 0.133 \\
    \midrule
    \textbf{QUEST$_{Full}$}                   & \textbf{0.736}$_{\pm\text{0.0022}}$ & \textbf{0.643}$_{\pm\text{0.0034}}$ & \textbf{0.212}$_{\pm\text{0.0024}}$ & \textbf{0.141}$_{\pm\text{0.0020}}$ \\
    \textbf{QUEST}$_{\text{w/o spectral init}}$ & \underline{0.735}$_{\pm\text{0.0028}}$ & \underline{0.640}$_{\pm\text{0.0039}}$ & \textbf{0.212}$_{\pm\text{0.0028}}$ & 0.138$_{\pm\text{0.0034}}$ \\
    \textbf{QUEST}$_{\text{w/o graph-reg}}$   & 0.734$_{\pm\text{0.0028}}$ & 0.637$_{\pm\text{0.0043}}$ & \underline{0.210}$_{\pm\text{0.0011}}$ & \underline{0.140}$_{\pm\text{0.0011}}$ \\
    \midrule
    $\Delta$ & \textcolor{teal}{$+1.2\%$} & \textcolor{teal}{$+1.1\%$} & \textcolor{teal}{$+2.4\%$} & \textcolor{teal}{$+6.0\%$} \\
    \bottomrule
  \end{tabular}
    \caption{Comparison of QUEST and baselines on NL27k and CN15k for Link Prediction. Best results are in \textbf{bold}, second-best \underline{underlined}. $\Delta$ denotes percentage improvement relative to ssCDL in \textcolor{teal}{green}.}
  \label{lp_result}

\end{table*}

\begin{table}[t]
  \centering
  \small
  \begin{tabular}{lrrr}
    \toprule
    Dataset & \#Entities & \#Relations & \#Quadruples \\
    \midrule
    NL27k & 27,221 & 404 & 175,412 \\
    CN15k & 15,000 & 36 & 241,158 \\
    \bottomrule
  \end{tabular}
  \caption{Statistics of the UKG datasets.}
  \label{tab:dataset}
\end{table}

\section{Experiments}
\label{sec:setup}

\subsection{Datasets}
\label{sec:datasets}

We evaluate QUEST on NL27k and CN15k~\cite{ukge2019}. NL27k is derived from the Never-Ending Language Learning (NELL) knowledge base~\cite{mitchell2018never}, which extracts relational facts from web pages and assigns confidence scores. CN15k is derived from ConceptNet~\cite{speerConceptNet55Open2017}, a multilingual semantic network encoding commonsense knowledge as weighted relations. Table~\ref{tab:dataset} reports dataset statistics.

\subsection{Baselines}
\label{sec:baselines}

We compare QUEST with prior UKG completion methods, including \textbf{UKGE}~\cite{ukge2019}, \textbf{PASSLEAF}~\cite{passleaf2021}, \textbf{UKGsE}~\cite{ukgse2022}, \textbf{BEUrRE}~\cite{beurre2021}, \textbf{UPGAT}~\cite{upgat2023}, and \textbf{ssCDL}~\cite{sscdl2025}, on confidence prediction and link prediction tasks. 

\begin{figure*}[t!]
  \centering
  \includegraphics[width=0.9\textwidth]{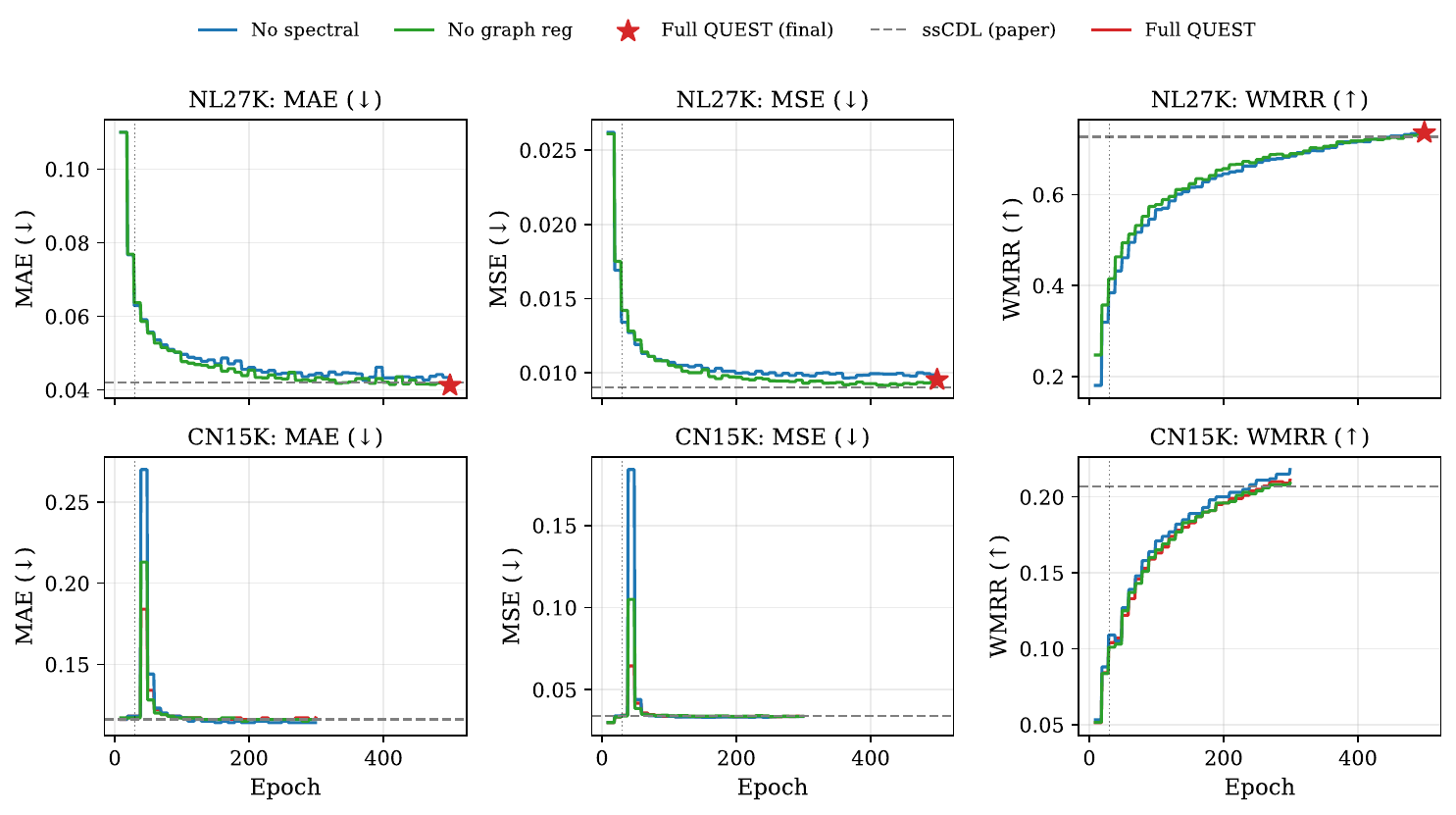}
  \caption{\textbf{Training dynamics.}  Validation MAE, MSE, and WMRR across epochs for the three QUEST variants on NL27k (top) and CN15k (bottom).  The vertical dotted line at epoch~30 marks the activation of ssCDL's PCDG self-training stage. The ssCDL reference is shown as a dashed grey line.}
  \label{fig:dynamics}
\end{figure*}

\subsection{Evaluation Protocol}
\label{sec:protocol}

We evaluate QUEST on two standard UKG completion tasks: \textbf{confidence prediction} and \textbf{link prediction}. The evaluation protocol follows the unKR~\cite{unkr} benchmark.

\textbf{Confidence Prediction.}
For confidence prediction, the model predicts a 101-bin confidence distribution $\hat{\mathbf{p}}$
over $[0, 1]$. The predicted confidence is the expected bin value:
\begin{equation}
  \tilde{s} \;=\; \sum_{b=0}^{B-1} \tfrac{b}{B-1}\,\hat{p}_b
\end{equation}
We normalize $\tilde{s}$ to $[0.1, 1.0]$ using a linear transformation:
\begin{equation}
  \hat{s} \;=\; \frac{\tilde{s}-\ell_{\sigma}}{u_{\sigma}-\ell_{\sigma}}\,0.9 + 0.1
  \label{eq:conf_norm}
\end{equation}
where $\ell_{\sigma} = \E_{b\sim\mathbf{p}_{0.1}}\bigl[\tfrac{b}{B-1}\bigr]$ and
$u_{\sigma} = \E_{b\sim\mathbf{p}_{1.0}}\bigl[\tfrac{b}{B-1}\bigr]$ are expected
bin values under Gaussians with bandwidth $\sigma=0.6$.
We report MSE and MAE between predicted and ground-truth values.

\textbf{Link Prediction.}
We perform tail-entity prediction in the filtered setting~\cite{ukge2019}. For each $(h,r,t,s)$, we score all candidate tails. We mask entities that form known true triples in the training, validation, or test set, except the ground-truth tail $t$, and compute the rank of $t$ using the standard double-argsort method. We report two metrics:
\begin{equation}
  \mathrm{WMRR} \;=\; \frac{\sum_i s_i\,\tfrac{1}{\mathrm{rank}_i}}{\sum_i s_i}
\end{equation}
\begin{equation}
  \mathrm{Hits@1} \;=\; \frac{1}{|\mathcal{Q}|}\sum_i \mathbb{I}[\mathrm{rank}_i = 1]
\end{equation}
where $\mathcal{Q}$ denotes the set of test queries.
WMRR aggregates reciprocal ranks weighted by ground-truth confidence, while Hits@1 measures the fraction of queries where the ground-truth tail ranks first. We use the default confidence filter of $0$, so all test triples contribute to LP evaluation.

\begin{figure*}[t]
  \centering
  \includegraphics[width=0.9\textwidth]{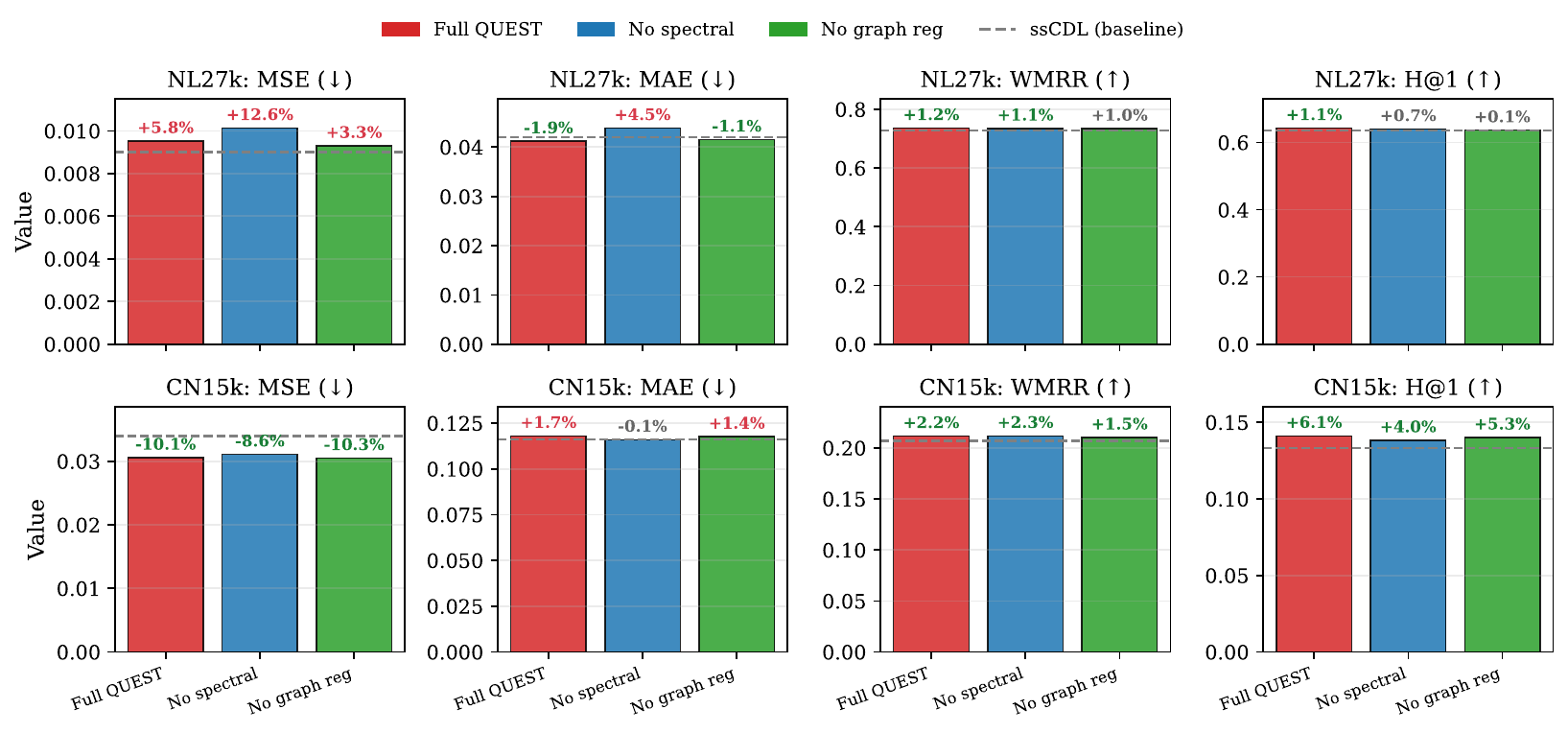}
  \caption{\textbf{Ablation study.}  Per-metric test-set values for
    full QUEST and the two single-component ablations across both
    datasets.  Dashed lines mark the ssCDL baseline.  Numbers above
    bars report $\Delta$ vs.\ ssCDL in percentage points
    (\textcolor{teal}{green}: improvement,
      \textcolor{red}{red}: regression,
  grey: $|\Delta|\!<\!1\%$).}
  \label{fig:ablation}
\end{figure*}

\subsection{Main Results}
\label{sec:main_results}

Tables~\ref{cp_result} and~\ref{lp_result} report performance on NL27k and CN15k for two tasks. QUEST$_\text{Full}$ achieves the best result on six of eight metric-dataset combinations and ties ssCDL on the remaining two (CN15k MSE and MAE), where a QUEST ablation still outperforms every non-QUEST baseline. The gains on CN15k consistently exceed those on NL27k: $+11.8\%$ MSE and $+6.0\%$ Hits@1 on CN15k, compared with $+2.4\%$ MAE and $+1.1\%$ Hits@1 on NL27k. This pattern is consistent with the confidence-weighted Laplacian construction in Section~Confidence-Weighted Graph Laplacian. CN15k contains roughly $2.5\times$ more edges per entity than NL27k, producing a denser Laplacian whose low-frequency eigenvectors carry a stronger community signal. The same spectral initialization therefore induces a stronger structural prior. On NL27k, the prior is weaker because the graph itself is sparser, while the pre-PCDG phase where $\Lsmooth$ reinforces that prior remains fixed at 30 of 500 epochs.

PASSLEAF and UPGAT are the closest conceptual baselines: PASSLEAF introduces semi-supervised triples, while UPGAT introduces graph attention, but both score triples as scalars. Neither matches ssCDL on NL27k link prediction, and only PASSLEAF$_\text{RotatE}$ approaches ssCDL on CN15k MSE (0.094 vs.\ 0.034). QUEST further reduces this error from 0.094 to 0.031, a $67\%$ reduction, without modifying the scoring function. This supports the claim in Section~QUEST that initialization geometry, rather than scorer expressiveness, is the main bottleneck.

\subsection{Training Dynamics}
\label{sec:dynamics}

Figure~\ref{fig:dynamics} shows that on NL27k, all three variants pass through epoch~30 smoothly. On CN15k, the two ablations spike to roughly $0.27$ MAE and $0.18$ MSE at epoch~30 and require $30$--$50$ epochs to recover, while QUEST$_\text{Full}$ shows no visible discontinuity. This matches the gradient-conflict mechanism in Section~PCDG-Aware Scheduling: when $\Lsmooth$ remains active after epoch~30, it pulls distant pseudo-corrupted entities together exactly when PCDG pushes them apart. The conflict scales with graph density, explaining why NL27k tolerates the same regularizer that destabilizes CN15k. A single scheduling rule, deactivating $\Lsmooth$ at the PCDG boundary, removes the spike on the dense graph without dataset-specific tuning.

Both ablations recover to within a few percent of QUEST$_\text{Full}$, so the final test scores understate the cost of leaving $\Lsmooth$ active. The practical issue appears during checkpoint selection: the MAE-minimizing checkpoint used by the ssCDL pipeline becomes biased toward pre-spike epochs whose representations have not yet incorporated PCDG supervision. Deactivating $\Lsmooth$ at epoch~30 removes this instability and improves reproducibility more than the small final-score gap itself.

\subsection{Ablation Study}
\label{sec:ablation}

Figure~\ref{fig:ablation} compares QUEST$_\text{Full}$,
QUEST$_\text{w/o spectral}$, and QUEST$_\text{w/o graph-reg}$ on both datasets. Removing spectral initialization causes a $4.5\%$ MAE regression on NL27k versus baseline, though it still achieves a $+1.1\%$ WMRR gain. This confirms that spectral initialization is the main contributor to regression quality on the sparse graph. Removing the graph regularizer still yields a $+3.3\%$ MSE improvement on NL27k, indicating that spectral initialization is sufficient for confidence prediction.

QUEST$_\text{w/o graph-reg}$ achieves the best CN15k MSE, outperforming QUEST$_\text{Full}$ on that metric alone. This is not an evidence against the schedule: the same configuration produces the spike in Figure~\ref{fig:dynamics}, so the lower MSE is reached through an unstable trajectory and remains sensitive to checkpoint selection. No ablation Pareto-dominates QUEST$_\text{Full}$ across the eight evaluation cells. The two components address different failure modes: initialization geometry and early-stage structural drift. Combining them is the only configuration that wins or ties on all cases avoiding the epoch-30 spike.

\section{Conclusion}
\label{sec:conclusion}

We introduced QUEST, a parameter-free extension of ssCDL for uncertain knowledge graph completion. QUEST initializes entities from the confidence-weighted Laplacian and uses an unbiased mini-batch Dirichlet regularizer during early training. It then deactivates the regularizer when PCDG begins, avoiding the gradient conflict observed on dense graphs. Across NL27k and CN15k, QUEST improves most confidence and link prediction metrics while removing the epoch-30 instability spike.

\paragraph{Limitations.}
QUEST is evaluated on two UKG benchmarks and inherits ssCDL's 101-bin confidence head and fixed PCDG schedule. Its one-time spectral step may become expensive on very large graphs, and its undirected Laplacian ignores relation direction and type.

\paragraph{Broader Impact.}
\label{sec:impact}
QUEST may improve systems that reason over uncertain facts, but UKGs can encode noise, bias, and coverage gaps. Predicted confidences should not be treated as ground truth in high-stakes settings without calibration checks and human oversight.

\bibliography{main}

\end{document}